\documentclass[runningheads]{llncs}

\usepackage[T1]{fontenc}
\usepackage{multirow} 
\usepackage{amssymb}
\newcommand{\cmark}{\checkmark}
\newcommand{\xmark}{$\times$}
\usepackage{graphicx}
\usepackage{epsfig}
\usepackage{amssymb}
\usepackage{makecell}
\usepackage{amsmath}
\usepackage{float}
\usepackage{subfigure}
\usepackage{authblk} % 加载 authblk 宏包
\usepackage{pifont} % 加载 pifont 宏包
\usepackage[colorlinks=true]{hyperref} 
\hypersetup{
    citecolor=blue
}
\usepackage{graphicx}
\usepackage{booktabs}
\begin{document}
\title{IRGNN: Efficient Invariant Radar Graph Neural Network for Radar Point Cloud Object Detection}
\titlerunning{IRGNN}
\author{
Xiao Guo\inst{1}
\and
Wanke Xia\inst{2}
\and
Lili Yang\inst{1}\thanks{Corresponding author: \url{llyang@cau.edu.cn}}
\and
Caicong Wu\inst{1}
}
\authorrunning{X. Guo et al.}
\institute{China Agricultural University
\and
Tsinghua University
}
\maketitle              % typeset the header of the contribution
\begin{abstract}
% Perception systems are an important part of autonomous driving. In recent years, thanks to the in-depth research on LiDAR (Light Detection and Ranging), the accuracy of the perception system has been continuously improved. However, the ability of LiDAR sensors to perceive in undesirable weather conditions is limited. Recently, radar point clouds have become popular in perception system researches due to the robustness against bad weathers and the availability in low illumination scenarios. However, the small amount of data contained in radar point clouds is a challenge for the current mainstream LiDAR-based perception methods. To address the challenge, we propose a novel method for object detection by inputting an invariant radar point cloud into the graph neural network. Firstly, radar point clouds are reconstructed into graph structures by using translation and rotation invariance methods, respectively, which is to overcome the weakness of radar data. Then, a graph neural network model is proposed by using improved message passing neural network structure. Finally, different heads use the output of the network for classification and detection tasks. The research and experiments are performed on the RadarScenes dataset. The results indicate that our method surpasses other existing methods based on the dataset, confirming the superiority of our approach. At the same time, we achieved a significant reduction in computational resource requirements.
Perception is a fundamental component of autonomous driving systems. While LiDAR-based methods have achieved remarkable progress in object detection, their reliability can degrade under adverse weather conditions. Radar point clouds provide a robust alternative due to their resilience to bad weather and low-illumination scenarios. However, radar point clouds are typically sparse, unordered, and less informative than LiDAR data, making it challenging to directly apply existing LiDAR-based perception methods.
To address these challenges, we propose \textbf{IRGNN}, an \underline{I}nvariant \underline{R}adar \underline{G}raph \underline{N}eural \underline{N}etwork for radar point cloud object detection. IRGNN first reconstructs radar point clouds into graph representations using translation- and rotation-invariant feature designs, enabling robust modeling of sparse radar measurements. It then employs an improved message passing neural network (MPNN) with residual connections and a virtual node layer to enhance local feature propagation and global context modeling. Finally, task-specific heads are applied to the learned graph representations for object classification and bounding box prediction.
Experimental results on the \textit{RadarScenes dataset} show that IRGNN outperforms existing radar-based object detection methods and achieves competitive performance. In addition, IRGNN significantly reduces computational cost and memory usage during inference, demonstrating its effectiveness and practical potential for efficient radar-based perception in autonomous driving.

\end{abstract}
\section{Introduction}

Accurate perception is essential for autonomous driving systems. Existing perception methods mainly rely on cameras and LiDAR sensors~\cite{liu2023real,alfred2023fully,hess2025splatad,carranza2022object,zimmer2023infradet3d}. Cameras provide rich texture and semantic information, but their performance is highly sensitive to illumination conditions and can degrade in overexposed or low-light environments~\cite{xia2025overall,xia2025improved,zhang2024gait}. LiDAR sensors provide accurate geometric measurements, but they are vulnerable to adverse weather conditions such as rain, snow, and fog, which may introduce noise, occlusion, and unreliable detections.
Millimeter-wave radar has become an important complementary sensor for robust autonomous driving perception~\cite{harlow2024new,cheng2026radarmp,zeng2026velocity,zhang2026dsfc,tian2026extended}. Compared with cameras and LiDAR, radar is less affected by bad weather and low illumination and can also directly measure target velocity, which provides useful motion information for dynamic driving scenarios. However, radar point clouds are usually sparse, noisy, unordered, and less informative than LiDAR data, which makes it difficult to directly apply LiDAR-based object detection methods to radar data.

Graph neural networks (GNNs) offer a natural way to model radar point clouds. By treating radar points as nodes and point-wise relationships as edges, GNNs can preserve the irregular structure of point clouds and capture local geometric dependencies. However, two challenges remain for graph-based radar perception. First, radar point cloud features are sensitive to coordinate transformations, which weakens the robustness of learned representations. Second, standard message passing networks may suffer from limited global context modeling and over-smoothing when processing sparse radar graphs.

To address these challenges, we propose \textbf{IRGNN}, an \underline{I}nvariant \underline{R}adar \underline{G}raph \underline{N}eural \underline{N}etwork for radar point cloud object detection. IRGNN first reconstructs radar point clouds into graph structures with translation- and rotation-invariant feature representations. Then, it introduces an improved message passing neural network (MPNN) with residual connections and a virtual node layer to enhance feature propagation and global context modeling. Finally, classification and detection heads are applied to the learned node representations for object recognition and bounding box prediction.
We evaluate IRGNN on the RadarScenes dataset~\cite{schumann2021radarscenes}, the representative large-scale high-resolution radar dataset that includes per-point annotations for moving instance tracking under versatile scenarios. Experimental results show that IRGNN outperforms comparable radar-based methods and achieves competitive performance.

In general, our main contributions can be summarized as follows:

(1) We propose IRGNN, an invariant radar graph neural network framework for radar point cloud object detection.

(2) We design translation- and rotation-invariant graph representations and improve MPNN with residual connections and a virtual node layer to enhance robustness, feature propagation, and global context modeling.

(3) Experimental results show that IRGNN outperforms comparable radar-based detection methods and reduces inference time and memory usage with post-processing optimization, demonstrating the practical potential for efficient radar-based perception in autonomous driving.
 % and achieves competitive performance against image-based baselines

\section{Related Works}
\subsection{Graph Neural Network in Point Cloud}
Early point cloud methods were largely inspired by CNN-based representations, where irregular points were converted into grids, voxels, or image-like structures~\cite{wang2019dynamic,chen2020hierarchical,yang2022graph}, but they may weaken the unordered and irregular nature of point clouds and introduce discretization costs. 
To better preserve geometric structures, GNN-based methods were introduced to directly model point clouds as graphs, where points are treated as nodes and local neighborhoods are represented as edges~\cite{shi2020point,wang2021point,ding2026memground}. 
These methods replace regular convolutions with graph convolutions or message passing, enabling more effective modeling of local spatial relationships. 
More recently, GNNs have been extended to radar point cloud perception, showing advantages in modeling sparse and unstructured radar measurements~\cite{svenningsson2021radar,fent2023radargnn,hunt2025ragnnarok,fang2024real}. However, radar point clouds remain challenging due to sparsity, noise, weak geometric information, and sensitivity to coordinate transformations, which motivates more robust and invariant graph representations.

\subsection{Invariance in Point Cloud}
Invariance is a key issue in point cloud learning, since point cloud features are sensitive to geometric transformations. Existing studies mainly address this problem from three directions. Transformation-based methods learn alignment or pose transformation modules to normalize point clouds before feature extraction, but they do not strictly guarantee invariance \cite{qi2017pointnet,you2018pvnet,yuan2018iterative,fei2024rotation}. Value-based methods construct invariant descriptors from geometric relations such as distances, angles, point-pair features, or Gram matrices, providing stricter invariance but relying heavily on the discriminative ability of handcrafted geometric descriptors \cite{deng2018ppfnet,zhang2019rotation,xu2021sgmnet}. LRF-based methods further establish local reference frames or local coordinate systems to represent points under a consistent local basis, improving rotation robustness and local geometric modeling, while their effectiveness depends on the stability and expressiveness of the constructed reference frame \cite{cao2020lfnet,chen2022devil,zhao2022rotation}. 
These studies motivate the use of invariant point-pair representations for robust radar point cloud modeling.

\section{Methods}

\begin{figure}[t]
  \centering
  \includegraphics[width=1\linewidth]{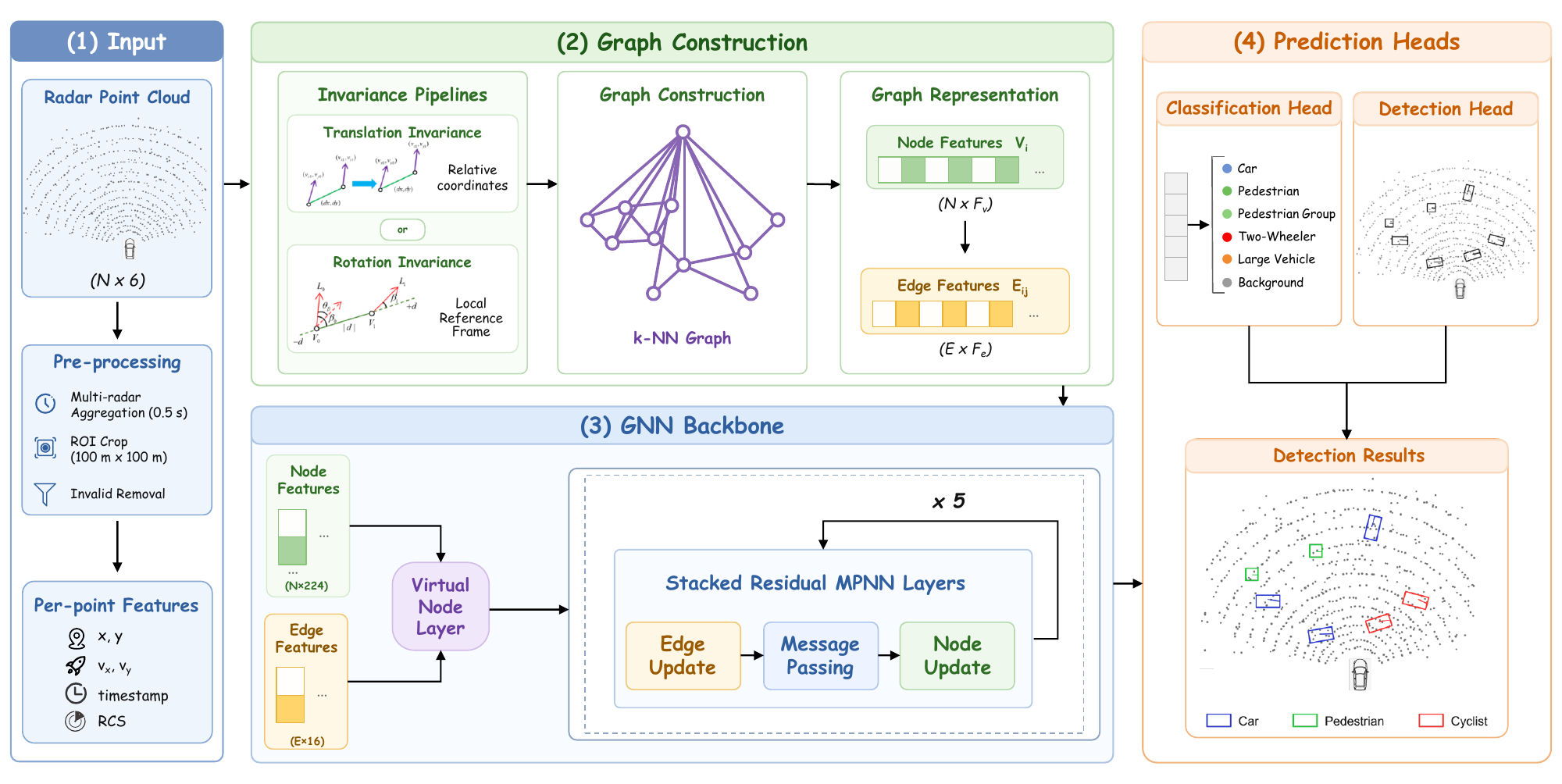}
  \caption{\textbf{Overview of IGRNN}, mainly comprising \textit{Graph Construction} and \textit{GNN}.}
  \label{fig:1}
\end{figure}

We propose \textbf{IRGNN}, an invariant radar graph neural network framework for radar point cloud object detection in autonomous driving scenarios (Figure \ref{fig:1}).

\subsection{Graph Construction}

\textbf{Data Preprocessing.} The original radar point cloud in RadarScenes~\cite{schumann2021radarscenes} contains 17-dimensional data, of which we use six types: $x$, $y$, $vx$, $vy$, timestamp, and RCS. 
Early data preparation follows RadarGNN~\cite{fent2023radargnn} and Scheiner's method~\cite{scheiner2021object}. 
RadarScenes applied multiple radars to gather data, and timing disparities can still be seen in the same frame of data from different radars. 
As a result, radar point clouds within 0.5 seconds are aggregated into a single frame and cropped to the area in front of the vehicle that is 100 by 100 meters. 
After that, points with invalid coordinates and velocities are removed.

\textbf{Transformation-Invariant Graph.} The graph is defined as $G(\mathcal{V}, \mathcal{E})$. The original point cloud is a finite set $\mathcal{P} = \{P_{1}, P_{2}, \dots, P_{n}\}$, where $P_{i}~(i \in n)$ is a vector containing the original point features of $F_{p}$ dimensions.
$P_{i}$ is mapped to $V_{i}~(i \in n)$, a vector containing the node features of $F_{v}$ dimensions. The final node set with features is $\mathcal{V} = \{V_{1}, V_{2}, \dots, V_{n}\}$. Edges $\mathcal{E}$ are computed by k-Nearest Neighbor (KNN). Formally, the edge between node $u$ and node $v$ is defined as:
\begin{equation}
E_{uv}=\left\{\left(V_{u}, V_{v}\right) \mid\left(V_{u}, V_{v}\right) \in \mathcal{V}^{2}, u \neq v\right\}
\end{equation}
where $V_{u}$ denotes node $u$ and $V_{v}$ denotes node $v$. Also, node $v$ with feature is:
\begin{equation}
V_{v}=KNN_{\sigma}\left(V_{u}\right)
\end{equation}
where $KNN_{\sigma}$ denotes KNN with $\sigma$ neighbors.
% In this paper, $\sigma=20$.
The edges carry point-to-point features between nodes with dimension of $F_{e}$.
To achieve transformation invariance, we apply two strategies: translation invariance and rotation invariance.

\begin{figure}[t]
\centering
\begin{minipage}{0.48\textwidth}
  \centering
  \includegraphics[width=\linewidth]{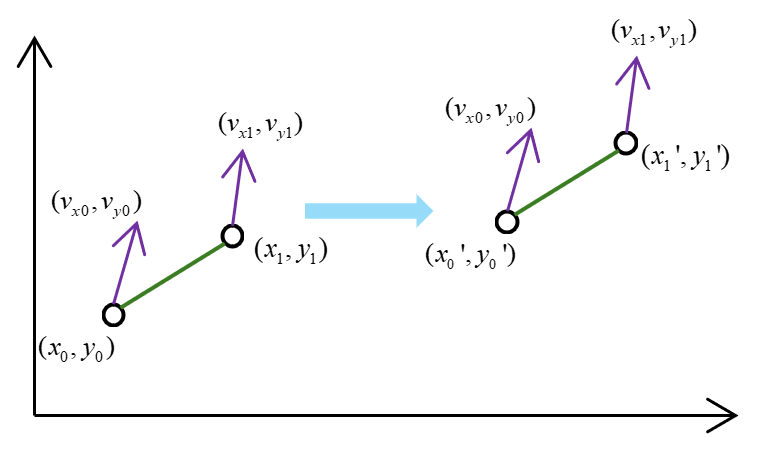}
  \centering (a)
\end{minipage}
\hfill
\begin{minipage}{0.48\textwidth}
  \centering
  \includegraphics[width=\linewidth]{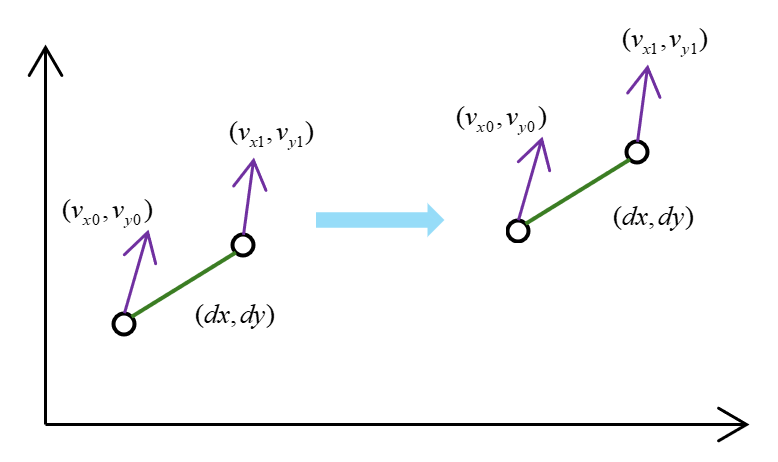} 
  \centering (b) 
\end{minipage}
\caption{Schematic diagram of translation invariance: (a) no invariance (position changes after translation); (b) translation invariance (relative position does not change).}
% (a) no invariance method is used, the position information has changed after translation; (b) a translation invariance method is used, the relative position information will not change.
\label{fig:2}
\end{figure}

\textbf{Translation Invariance.} In a graph without invariance implementation, all six-dimensional features are encoded into point features, while edge features remain empty. However, $x$ and $y$ cannot be explicitly encoded into point features as translation-invariant properties because their values change as the point is translated. As a result, they are converted to relative coordinates $(dx, dy)$ between the two nodes and become edge features between the two points, as illustrated in Figure \ref{fig:2}. In addition, the timestamps in the point features are converted to time indexes, and the point connectivity information $c$ is included.

\begin{figure}[t]
\centering
\begin{minipage}{0.48\textwidth}
  \centering
  \includegraphics[width=\linewidth]{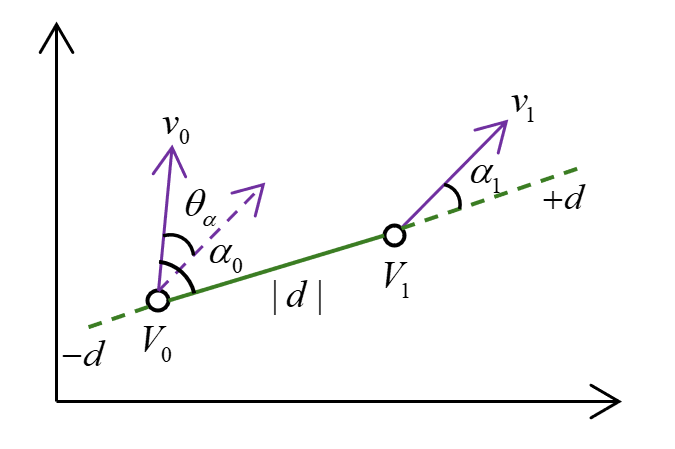}
  \centering (a)
\end{minipage}
\hfill
\begin{minipage}{0.48\textwidth}
  \centering
  \includegraphics[width=\linewidth]{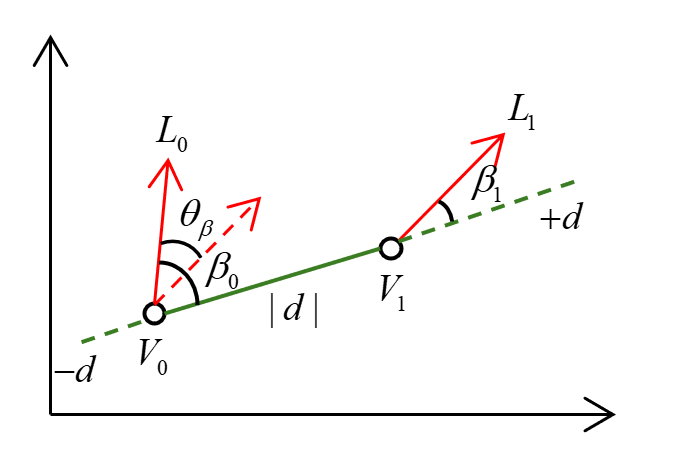} 
  \centering (b) 
\end{minipage}
\caption{Schematic diagram of rotation invariance: (a) RadarGNN's~\cite{fent2023radargnn}; (b) Our EPPF.}
\label{fig:3}
\end{figure}

\textbf{Rotation Invariance.} When rotating point clouds, the velocities and relative positions also change. RadarGNN~\cite{fent2023radargnn} uses a set of point pair features to solve the rotation problem, as shown in Figure \ref{fig:3} (a). These features include the distance $|d|$ between nodes, the relative angle $\theta_{a}$ of the velocities, and the relative angles $a_{1}$ and $a_{2}$ between the velocities of two nodes and the edges.

\textbf{Enhanced Point Pair Feature.}
While RadarGNN~\cite{fent2023radargnn} guarantees rotation invariance, it falls short in capturing local geometric features. 
To address this issue, we introduce an enhanced point pair feature method (EPPF) based on a local reference frame (LRF) to improve the expressiveness and discriminability of local features, as shown in Figure \ref{fig:3} (b).
Specifically, for a given current node and its set of neighborhood points, a stable and unique LRF must first be established. 
The construction of LRF relies on the spatial distribution characteristics of the neighborhood points.
By constructing a weighted covariance matrix for neighborhood points and performing eigen-decomposition on the matrix, a set of eigenvalues and corresponding eigenvectors can be obtained. 
The eigenvector corresponding to the maximum eigenvalue is designated as the reference axis of the local reference coordinate system. 
Based on this local reference axis (LRA), the computation of point-to-point features remains consistent under different rotation conditions. 
Concretely, the calculation process is as follows. 
First, we obtain the displacement vector of the $u$-th neighbor $a_{i,u}$ relative to node $a_{i}$:
\begin{equation}
b_{u}=a_{i, u}-a_{i}\left(a_{i} \in \mathbb{R}^{2}, KNN_{a}=\left\{a_{i, 1}, a_{i, 2}, \dots, a_{i, a}\right\}\right)
\end{equation}
% The neighbor points are obtained by KNN.
Then, we compute the distance weight of the $u$-th neighbor as:
\begin{equation}
w_{u}=\frac{\left\| b_{u}\right\| ^{-1}}{\sum_{n=1}^{a}\left\| b_{n}\right\| ^{-1}}
\end{equation}
The closer the $u$-th neighbor point is to node $a_i$, the heavier the weight $w_{u}$ is. 
Now, the weighted covariance matrix $M_{i}$ is defined as:
\begin{equation}
M_{i}=\sum_{n=1}^{a} w_{n} b_{n} b_{n}^{\top}
\end{equation}
EPPF consists of relative angles $\beta_{0}$ and $\beta_{1}$ between LRA and the edge, and the angle $\theta_{\beta}$ of LRA. 
Finally, EPPF is computed as:
\begin{equation}
\text{EPPF}\left(V_{0,1}\right)=\left[|d|, a_{1}, a_{2}, \theta_{\alpha}, \beta_{1}, \beta_{2}, \theta_{\beta}\right]
\end{equation}
\begin{equation}
\left\{
\begin{array}{l}
|d|=\left\| V_{1}-V_{0}\right\|, \\
a_{0}=\angle^{ac}\left(v_{0}, d\right),
a_{1}=\angle^{ac}\left(v_{1}, d\right), 
\theta_{\alpha}=\angle^{ac}\left(v_{0}, v_{1}\right), \\
\beta_{0}=\angle^{ac}\left(L_{0}, d\right), 
\beta_{1}=\angle^{ac}\left(L_{1}, d\right), 
\theta_{\beta}=\angle^{ac}\left(L_{0}, L_{1}\right)
\end{array}
\right.
\end{equation}
where $\angle^{ac}$ represents an acute angle.

\subsection{Graph Neural Network}
\textbf{Feature Embemdding.}
Node and edge features are initially sent into the feature embedding layer, which employs a multi-layer perceptron (MLP) to expand the dimensions of features. 

\textbf{Virtual Node Layer.} 
We design a virtual node layer right after the feature embedding layer to enhance GNN's capacity to learn global information. 
% Gilmer~\cite{gilmer2017neural} was the first to suggest this concept, which adds a virtual node to every graph and creates virtual edges connecting it to every other node in the graph. 
The virtual node feature vector for the $i$-th graph $G_{i}(V, E)$ is:
\begin{equation}
VN_{i} \in \mathbb{R}^{F_{E}}
\end{equation}
where $F_{E}$ is the feature dimension of the virtual node.
Here, $F_{E}$ aligns with the first-layer node features. 
% In theory, this feature dimension might be set to any value, allowing for more flexible model capacity. 
% However, if the dimension does not match the first-layer node features, extra linear projections are needed at the input stage to ensure dimensional alignment, which increases model complexity and introduces new parameters. 
% To minimize this extra overhead, this study directly sets $F_{E}$ to be consistent with the first-layer node features.
In the forward propagation, virtual node features are concatenated with the embedded node feature vectors to form the input representation of the first-layer node features, which is formally as:
\begin{equation}
\tilde{X}_{i}(1)=X_{i}(1)+VN_{i}
\end{equation}
where $X_{i}(1)$ denotes the original node features of first layer for the $i$-th graph.
% This design not only ensures the consistency of the input dimension but also can inject global context information at the initial stage of the model. 

\textbf{Message Passing Neural Network.} 
MPNN updates node representations by passing messages along graph edges. For each target node $V_i$, a message is first generated from its current node feature $F_i^V$, the neighbor node feature $F_j^V$, and the edge feature $F_{ij}^E$:
\begin{equation}
M_{ij}=\phi_e\left(F_i^V,F_j^V,F_{ij}^E\right),
\end{equation}
where $\phi_e$ is the generate function by MLP. Then, all messages from the neighborhood $N(i)$ are aggregated to obtain the neighborhood representation:
\begin{equation}
M_i=\bigoplus_{j\in N(i)}M_{ij}.
\end{equation}
Finally, the target node feature is updated by combining its original feature with the aggregated message:
\begin{equation}
F_i^V=\phi_u\left(F_i^V,M_i\right),
\end{equation}
where $\phi_u$ is the update function by MLP. Through this process, MPNN captures local structural information from neighboring radar points.

% Message Passing Neural Network (MPNN) generates messages on the graph edge by edge. 
% Messages are embedded by the target node features, neighbor node features, and edge features. 
% After that, all messages in the neighborhood are aggregated to the target node, and then an update function is used to update the current node features and the aggregated neighborhood messages to a new node feature representation. The calculation process is as follows.
% Given the target node features $F^{V}_{i}$ and neighbor node features $F^{V}_{j}$, for each edge feature $F^{E}_{ij}$ pointing from the neighbor to the target node, the message $M_{ij}$ is defined as described in Formula (10), where $\phi_{e}$ is the message function implemented by a set of MLP. Subsequently, all incoming messages within the neighborhood $N(i)$ are aggregated dimensionally to obtain $M_{i}$. Finally, using the update function $\phi_{u}$ implemented by another set of MLPs, the current target node features are fused with the aggregated neighborhood messages to obtain the new node feature representation.
% \begin{equation}
% M_{i j}=\phi_{e}\left(F_{i}^{V}, F_{j}^{V}, F_{i j}^{E}\right)
% \end{equation}
% \begin{equation}
% M_{i}=\bigoplus_{j \in N(i)} M_{i j}
% \end{equation}
% \begin{equation}
% F_{i}^{V}=\phi_{u}\left(F^{V}_{i}, M_{i}\right)
% \end{equation}

\textbf{Residual Block.} 
Repeated neighborhood aggregation in MPNN may smooth node representations, causing nodes within the same connected component to become indistinguishable as the network goes deeper. 
To alleviate this issue, we add a residual block after each MPNN layer. Given the MPNN output $\tilde{F}_i^l$ at the $l$-th layer and the previous-layer feature $F_i^{l-1}$, the residual output is defined as:
\begin{equation}
\hat{F}_{i}^{l}=
\begin{cases}
\tilde{F}_{i}^{l}+\gamma F^{l-1}_{i}, & D_{l-1}=D_{l} \\
\tilde{F}_{i}^{l}+\gamma p(l) F^{l-1}_{i}, & D_{l-1} \neq D_{l}
\end{cases}
\end{equation}
where $\gamma$ is the residual scaling factor, and $p(l)$ is a linear projection used only when the feature dimensions of adjacent layers are different. 
% By adding the previous-layer representation back to the current output, the residual block preserves unsmoothed information and improves gradient propagation. 
% The resulting feature 
$\hat{F}_i^l$ is then passed through batch normalization and ReLU to obtain the final layer output $F_i^l$.

\textbf{Loss Function.} 
We use a combined loss function consisting of weighted classification loss $L_{c l s}$ and bounding box loss $L_{b b o x}$:
\begin{equation}
L=\lambda_{c} L_{c l s}+\lambda_{b} L_{b b o x}
\end{equation}
where $\lambda_{c}$ and $\lambda_{b}$ represent the weights of $L_{c l s}$ and $L_{b b o x}$, respectively. 
The classification loss uses weighted cross entropy loss:
\begin{equation}
L_{c l s}=-\frac{1}{N} \sum_{i=1}^{N} \gamma_{c} \log \left(q_{i, c}\right)
\end{equation}
where $N$ is the total number of samples in a batch, $c$ is a class in the total category $C$, $\gamma_{c}$ represents the weight of class $c$, and $q_{i,c}$ represents the predicted probability that the $i$-th sample belongs to class $c$.
The bounding box loss uses Huber loss, which is a loss function between L1 and L2 loss:
\begin{equation}
\Delta_{z}=\hat{z}-z
\end{equation}
\begin{equation}
L_{b b o x}=
\begin{cases}
\frac{1}{2} \Delta_{z}^{2}, & \left|\Delta_{z}\right| \leq \delta \\
\delta\left(\left|\Delta_{z}\right|-\frac{1}{2} \delta\right), & \left|\Delta_{z}\right|>\delta
\end{cases}
\end{equation}
where $\hat{z}$ is the prediction, $z$ is the ground truth, and $\Delta_{z}$ represents the error between the prediction and the ground truth. $\delta$ is the smoothing threshold used to determine the boundary between L1 and L2 losses.

\textbf{Post Processing.} 
The classification head maps the final node features to class scores, while the bounding box head predicts candidate boxes for each node using a two-layer MLP with ReLU. During post processing, candidates classified as background or with high background confidence are first removed. The remaining boxes are then converted from the relative coordinate system of the invariant representation to the absolute coordinate system. Next, non-maximum suppression (NMS) is applied within each frame to remove duplicate detections. Finally, a confidence threshold is used to filter low-score boxes and obtain the final detection results.

\section{Experiments}
\subsection{Experiment Setup}
% IRGNN is implemented in Python 3.10, PyTorch 2.4, the PyTorch geometry library 2.6.1, and CUDA 11.8. 
During graph construction, we employ two invariance configurations in Table \ref{tab:1}. 
In $KNN_{\sigma}$, we set $\sigma$ to 20. In the residual block, we set $\gamma$ to 0.3.
% we test various MPNNs to determine the optimal feature transfer method. 
% Various virtual node strategies were tested to explore the impact of global information on GNNs. 
During GNN, node features are increased to 224 dimensions, while edge features are increased to 16.
IRGNN is trained for 40 epochs on an NVIDIA RTX4090 24 GB GPU with a learning rate of 0.001 and a batch size of 5. To avoid overfitting, we set the learning rate decay parameter to 0.95, and employ an early stopping strategy.
In the loss function, we set $\lambda_{c}$, $\lambda_{b}$ and $\delta$ to 1.0, 0.5 and 1.0, respectively.
We compare IRGNN with baseline RadarGNN~\cite{fent2023radargnn}, pseudo-image-based PointPillars~\cite{lang2019pointpillars}, point-set-based PointNet++~\cite{qi2017pointnet++}, and grid-based DOG~\cite{ronecker2024dynamic} on the RadarScenes dataset~\cite{schumann2021radarscenes}. All methods are evaluated under an $\text{IoU}$ of 0.3. 

\begin{table}[htbp]
\centering
\caption{Node and edge feature for translation invariance and rotation invariance.}
% , where the elements contained in EPPF are shown in Formula 7
\label{tab:1}
\setlength{\tabcolsep}{12pt}
\begin{tabular}{lcc}
\toprule
Invariance & Node Feature & Edge Feature \\
\midrule
Translation & $v_x,v_y,rcs,t_{idx},c$ & $dx,dy$ \\
Rotation & $\left|v\right|,rcs,t_{idx},c$ & EPPF \\
\bottomrule
\end{tabular}
\end{table}

\subsection{Evaluation Metrics}
We apply average precision (AP) and mean average precision (mAP) to evaluate the performance of object detection. 
Precision and Recall are computed as:
\begin{equation}
Pr(i)=\frac{\sum TP(i)}{\sum TP(i)+\sum FP(i)}, \quad Re(i)=\frac{\sum TP(i)}{G}
\end{equation}
where $TP(i)$ and $FP(i)$ represent the number of true positives and false positives up to the $i$-th prediction, respectively. $G$ represents the total number of ground truth for the category. 
Then, the AP for each class and mAP are calculated as:
\begin{equation}
AP_{c}(\tau)=\sum \Delta Re(j, j-1) \cdot Pr_{smooth}(j), \quad mAP(\tau)=\frac{1}{C} \sum_{c=1}^{C} AP_{c}(\tau)
\end{equation}
where $c$ represents a class in the class set $C$, $\Delta Re$ represents the difference in recall between two adjacent sampling points, $Pr_{smooth}$ represents the accuracy after non-increasing smoothing, and $\tau$ is the IoU threshold. 
% Finally, the arithmetic mean of the AP of all classes is obtained to get mAP.

\subsection{Experimental Results}
\textbf{Comparison Experiments.} 
As shown in Table~\ref{tab:2}, IRGNN achieves the best overall performance with a mAP of 0.591, outperforming RadarGNN by 0.024 and showing clear gains over PointNet++, PointPillars, and DOG. The results indicate that radar point cloud detection benefits more from structure-preserving graph modeling than from grid-based, pseudo-image-based, or pure point-set representations.
IRGNN also performs best on pedestrian group, two-wheeler, and large vehicle classes, indicating that point-pair relations and global context help capture object structures with diverse shapes and spatial distributions. 
In summary, IRGNN achieves a more balanced performance across categories and shows stronger robustness for radar point cloud detection.

\begin{table}[htbp]
\centering
\caption{Comparison results of mAP and class-wise AP by various detection methods.}
% RadarGNN and our method both use translation invariance settings
\label{tab:2}
\setlength{\tabcolsep}{4pt}
\begin{tabular}{lcccccc}
\toprule
% large vehicle
% pedestrian group
% two-wheeler
\multirow{2}{*}{Model} 
& \multirow{2}{*}{mAP} 
& \multirow{2}{*}{Car} 
& \multirow{2}{*}{Pedestrian} 
& \multirow{2}{*}{\makecell{Pedestrian\\Group}} 
& \multirow{2}{*}{\makecell{Two-\\Wheeler}} 
& \multirow{2}{*}{\makecell{Large\\Vehicle}} \\
& & & & & & \\ % 空行保持高度
\midrule
DOG~\cite{ronecker2024dynamic} & 0.327 & 0.387 & 0.306 & 0.345 & 0.357 & 0.241 \\
PointPillars~\cite{lang2019pointpillars} & 0.392 & 0.148 & 0.148 & 0.279 & 0.405 & 0.549 \\
PointNet++~\cite{qi2017pointnet++} & 0.459 & 0.531 & \textbf{0.336} & 0.515 & 0.522 & 0.390 \\
RadarGNN~\cite{fent2023radargnn} & 0.567 & \textbf{0.709} & 0.277 & 0.560 & 0.588 & 0.701 \\
% YOLOv3 & 56.9 & 69.1 & \textbf{37.0} & 57.8 & 55.9 & 64.8 \\
IRGNN (Ours) & \textbf{0.591} & 0.696 & 0.286 & \textbf{0.600} & \textbf{0.639} & \textbf{0.735} \\
\bottomrule
\end{tabular}
\end{table}

\begin{figure}[t]
  \centering
  \includegraphics[width=0.8\linewidth]{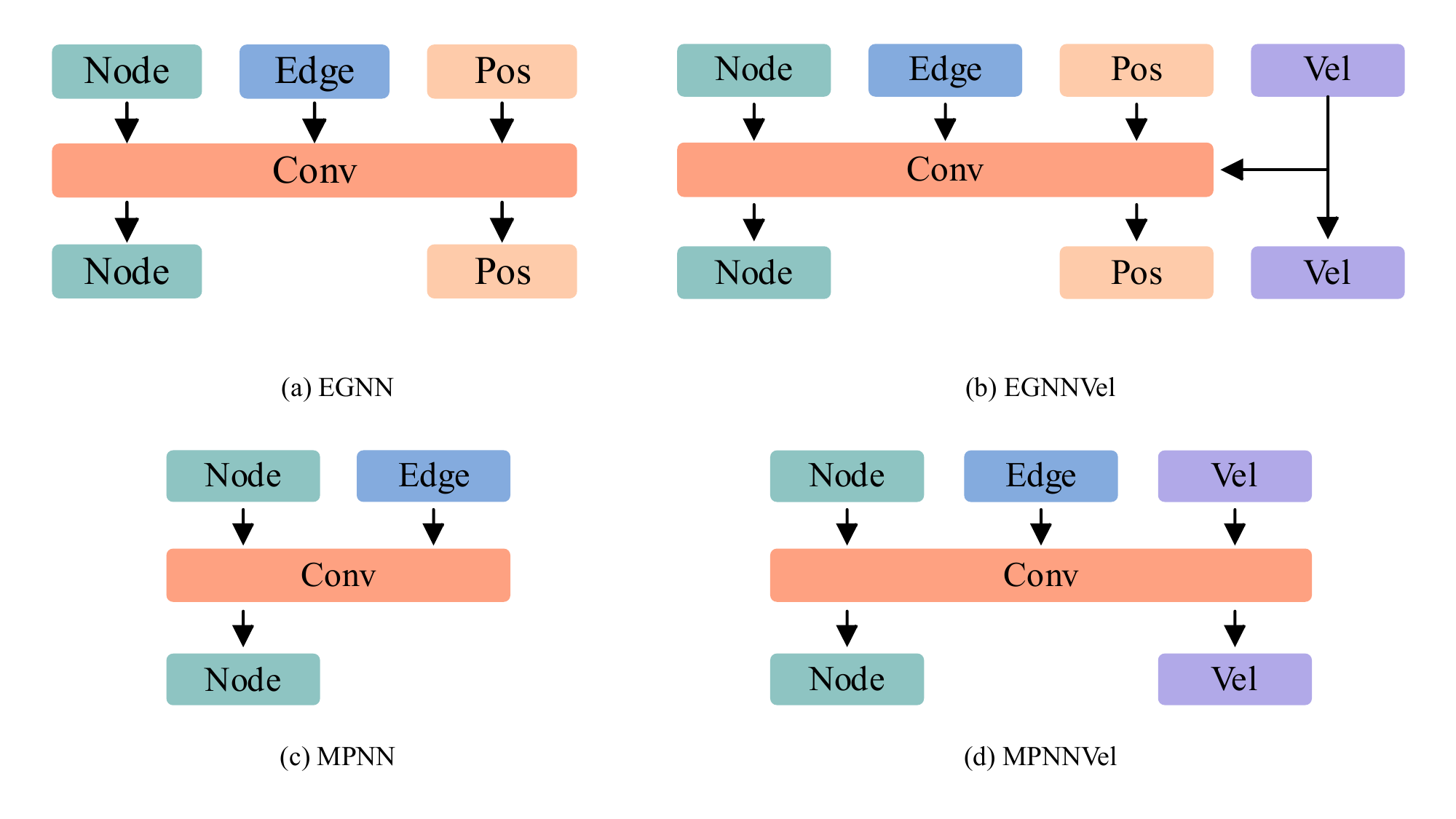}
  \caption{Schematic diagram of different convolutional layers.}
  \label{fig:5}
\end{figure}

\textbf{Convolutional Layer Experiments.}
To analyze the influence of different graph convolution designs on radar point cloud detection, we compare four representative layers shown in Figure \ref{fig:5}: EGNN~\cite{satorras2021n}, EGNNVel~\cite{satorras2021n}, MPNN, and MPNNVel. 
EGNN is selected as the geometric equivariant baseline, since it jointly updates node features and coordinates while preserving equivariance to translation, rotation, permutation, and reflection. 
EGNNVel further introduces velocity information into the coordinate update process, allowing motion cues to participate in geometric feature propagation. 
In contrast, MPNN follows a feature-centric message passing paradigm, where messages are generated from node and edge features, aggregated from neighboring nodes, and used to update node representations without explicitly modifying positions or velocities. 
To examine whether explicit motion modeling benefits MPNN, we additionally design MPNNVel, which removes velocity vectors from node features and instead encodes relative velocity differences into edge messages, followed by additive velocity-state regression. 

\begin{table}[htbp]
\centering
\caption{Results of mAP and class-wise AP with different convolutional layers.}
\label{tab:3}
\setlength{\tabcolsep}{6pt}
\begin{tabular}{lcccccc}
\toprule
\multirow{2}{*}{Layer} 
& \multirow{2}{*}{mAP} 
& \multirow{2}{*}{Car} 
& \multirow{2}{*}{Pedestrian} 
& \multirow{2}{*}{\makecell{Pedestrian\\Group}} 
& \multirow{2}{*}{\makecell{Two-\\Wheeler}} 
& \multirow{2}{*}{\makecell{Large\\Vehicle}} \\
& & & & & & \\ % 空行保持高度
\midrule
EGNN~\cite{satorras2021n} & 0.531 & 0.678 & 0.251 & 0.495 & 0.583 & 0.646 \\
EGNNVel~\cite{satorras2021n} & 0.552 & \textbf{0.723} & \textbf{0.346} & 0.540 & 0.523 & 0.630 \\
MPNN & \textbf{0.567} & 0.709 & 0.277 & \textbf{0.560} & \textbf{0.588} & \textbf{0.701} \\
MPNNVel & 0.510 & 0.693 & 0.218 & 0.447 & 0.514 & 0.680 \\
\bottomrule
\end{tabular}
\end{table}

% As shown in Table \ref{tab:3}, MPNN still has the highest mAP of 0.567. This may be because the node scale of the single graph built based on the RadarScenes dataset is significantly larger than the typical scene in the original EGNN. Therefore, we compared the training loss in various cases. The loss of EGNN and EGNNVel is higher than that of MPNN, which suggests that EGNN and its variants may have more difficulty generalizing on larger and denser graphs. As for our proposed MPNNVel, no performance improvement was observed.
As shown in Table~\ref{tab:3}, MPNN achieves the best overall mAP of 0.567, outperforming EGNN and EGNNVel despite their explicit geometric equivariance. 
This suggests that, for sparse and noisy radar point clouds, stable feature-level message passing is more effective than repeatedly updating coordinates or motion states. 
EGNNVel obtains the best AP on Car and Pedestrian, indicating that velocity cues can help objects with relatively clear motion or reflection patterns. 
However, its lower performance on Pedestrian Group, Two-Wheeler, and Large Vehicle shows that velocity-aware coordinate updates do not consistently generalize across categories. 
MPNNVel further decreases mAP to 0.510, suggesting that explicitly regressing velocity states may amplify radar noise rather than provide robust motion information.

\begin{figure}[tb]
  \centering
  \includegraphics[width=0.8\linewidth]{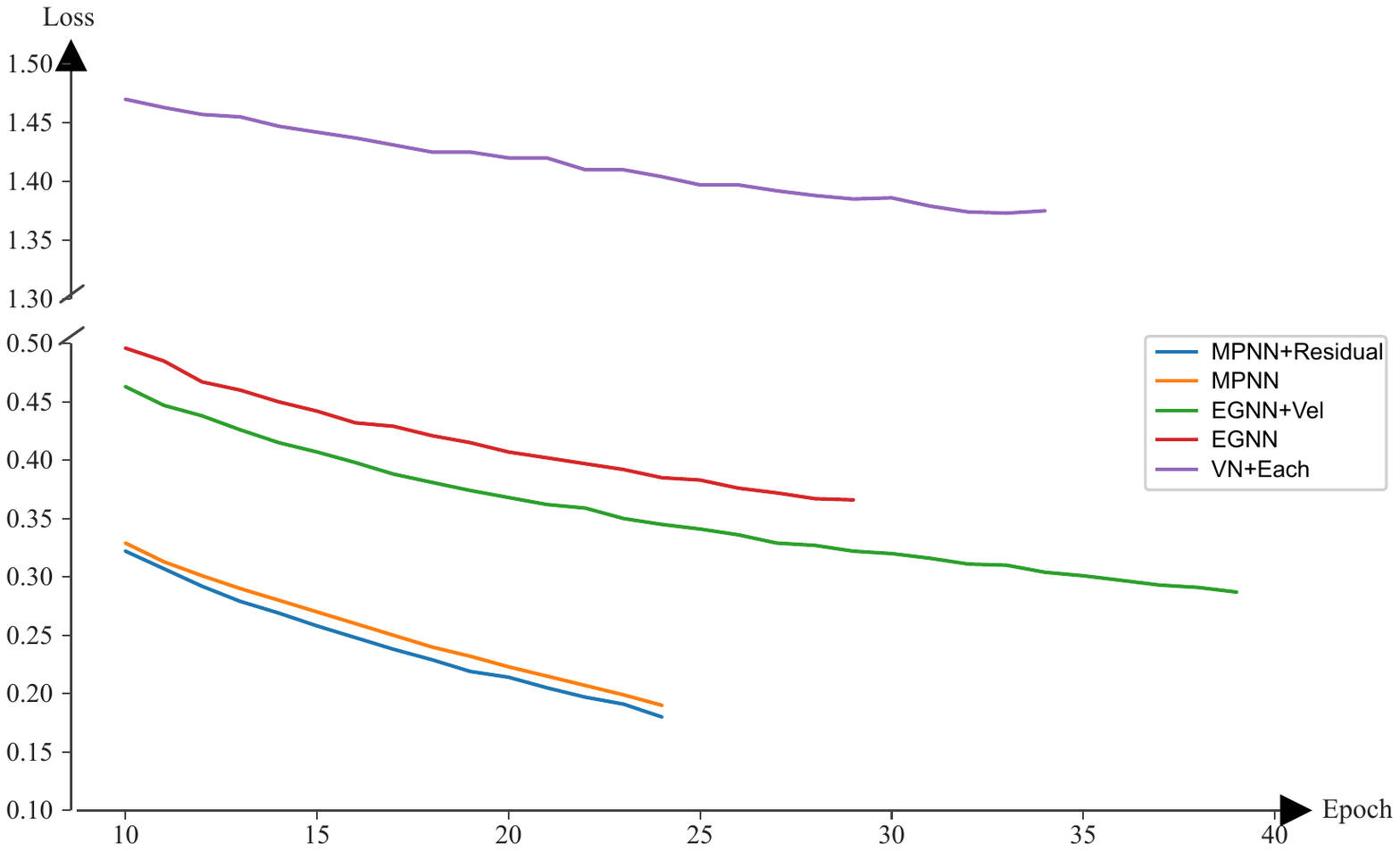}
  \caption{Training loss with different convolutional layers and improvements. All training processes have triggered the early stopping.}
  \label{fig:6}
\end{figure}

\textbf{Rotation Invariance Experiments.}
To evaluate the effectiveness of different rotation-invariant feature designs, we compare IRGNN with RadarGNN~\cite{fent2023radargnn} and RIConv++~\cite{zhang2022riconv++}. 
RadarGNN is based on point-pair geometric relations, while RIConv++ is a LRF-based method that constructs LRA and designs informative rotation-invariant features (IRIF) to enhance local geometric discrimination. 
For a fair comparison, we only replace the edge representation in the rotation-invariant pipeline and keep all other feature settings consistent.
Specifically, EPPF is replaced with IRIF when evaluating RIConv++.

% We conducted experiments using the rotation invariance pipeline described before and incorporated RIConv++~\cite{zhang2022riconv++}. This method builds on the idea of a local reference frame and designs a local reference axis to construct rotation-invariant local coordinates. Furthermore, it proposes Informative Rotation-Invariant Features (IRIF). In addition to characterizing the relationship between a target point and its neighborhood, it also explicitly encodes relationships between point pairs within the neighborhood, improving feature discrimination. In our implementation, we follow this principle, replacing the EPPF in Table 1 with IRIF, while keeping all other feature settings unchanged for consistency.

\begin{table}[htbp]
\centering
\caption{Results of mAP and class-wise AP with different rotation invariance pipelines.}
\label{tab:4}
\setlength{\tabcolsep}{4pt}
\begin{tabular}{lcccccc}
\toprule
\multirow{2}{*}{Model} 
& \multirow{2}{*}{mAP} 
& \multirow{2}{*}{Car} 
& \multirow{2}{*}{Pedestrian} 
& \multirow{2}{*}{\makecell{Pedestrian\\Group}} 
& \multirow{2}{*}{\makecell{Two-\\Wheeler}} 
& \multirow{2}{*}{\makecell{Large\\Vehicle}} \\
& & & & & & \\ % 空行保持高度
\midrule
RadarGNN~\cite{fent2023radargnn} & 0.229 & 0.229 & 0.118 & 0.267 & 0.337 & \textbf{0.196} \\
RICONV++~\cite{zhang2022riconv++} & 0.230 & 0.258 & 0.102 & 0.256 & \textbf{0.339} & 0.193 \\
IRGNN (Ours) & \textbf{0.238} & \textbf{0.259} & \textbf{0.133} & \textbf{0.332} & 0.277 & 0.190 \\
\bottomrule
\end{tabular}
\end{table}

As shown in Table~\ref{tab:4}, our IRGNN achieves the highest mAP of 0.238, slightly outperforming RadarGNN and RIConv++. 
The improvement mainly comes from Pedestrian and Pedestrian Group, indicating that the proposed EPPF can better capture local point-pair geometry in sparse radar regions. 
However, the overall gain is limited, and \textbf{all rotation-invariant pipelines perform much worse than the translation-invariant setting}. 
Although rotation-invariant descriptors improve robustness to pose changes, they may also remove useful orientation-related cues for bounding box prediction. 
% In radar point clouds, where observations are already sparse and noisy, such loss of directional information can weaken feature discriminability. 
% Therefore, the results show that our EPPF improves rotation-invariant modeling, but also reveal the inherent trade-off between strict rotation invariance and detection accuracy.

% As shown in Table 4, the overall performance of RIConv++ is close to that of the baseline method; our approach surpasses both in mAP, but the lead is limited. Furthermore, we observe that the mAP of the rotation invariance pipeline is still lower than that of the translation invariance pipeline. This phenomenon aligns with Zhou's observation that features constructed based on rotation invariance pipelines often lack discriminability compared to translation invariance pipelines, resulting in performance gaps on downstream tasks. These results not only validate the superiority of our approach over similar rotation-invariant methods and baselines, but also demonstrate that there is still room for improvement when dealing with complex box definitions.

\subsection{Ablation Study}
We conduct ablation experiments on residual blocks and virtual node layer (Table \ref{tab:5}).
All experiments are performed on the translation invariance pipeline.

\begin{table}[t]
\centering
\caption{Results of mAP and class-wise AP in ablation studies.}
\label{tab:5}
\setlength{\tabcolsep}{3pt}
\begin{tabular}{cccccccc}
\toprule
\multirow{2}{*}{\makecell{Residual\\Blocks}} 
& \multirow{2}{*}{\makecell{Virtual Node\\Layer}} 
& \multirow{2}{*}{mAP} 
& \multirow{2}{*}{Car} 
& \multirow{2}{*}{Pedestrian} 
& \multirow{2}{*}{\makecell{Pedestrian\\Group}} 
& \multirow{2}{*}{\makecell{Two-\\Wheeler}} 
& \multirow{2}{*}{\makecell{Large\\Vehicle}} \\
& & & & & & & \\ % 空行保持高度
% RB & VNL & mAP & Car & Pedestrian & Pedestrian Group & Two-Wheeler & Large Vehicle \\
\midrule
\xmark & \xmark& 0.567 & 0.709 & 0.277 & 0.560 & 0.588 & 0.701 \\
\cmark & \xmark & 0.578 & \textbf{0.727} & 0.242 & 0.570 & 0.638 & 0.711 \\
\cmark & \cmark & \textbf{0.591} & 0.696 & \textbf{0.286} & \textbf{0.600} & \textbf{0.639} & \textbf{0.735} \\
\bottomrule
\end{tabular}
\end{table}

\textbf{Ablation of Residual Blocks.} 
% The second row of Table \ref{tab:5} shows the performance improvement achieved by introducing residual blocks into the baseline model. Specifically, the residual structure improves the model's mAP by 0.9\% compared to the baseline method. This result demonstrates that residual blocks play a positive role in preventing feature over-smoothing that occurs with deeper graph neural network layers, effectively reducing the risk of representation degradation and further enhancing the detection head's ability to detect and localize objects.
% In addition, we also compared the convergence of training loss during training. As shown in Fig. 6, the orange curve corresponds to the training loss of the baseline method, while the blue curve shows the change in training loss after only the residual block is added under the same settings. It can be clearly seen that the residual block leads to a greater overall reduction in training loss and faster convergence. This phenomenon demonstrates that residual connections provide a shorter and more stable path for gradients in backpropagation, thereby reducing instability during training.
As shown in Table~\ref{tab:5}, adding residual blocks improves the mAP from 0.567 to 0.578, with clear gains on Car, Pedestrian Group, Two-Wheeler, and Large Vehicle. 
This indicates that residual connections help preserve earlier-layer node information during repeated message passing, thereby alleviating representation degradation caused by over-smoothing. 
The improvement is especially evident on categories with stronger or more spatially extended radar reflections, such as Two-Wheeler and Large Vehicle, where stable feature propagation is important for maintaining object-level structures. 
Although the AP for Pedestrian decreases, this may be due to the extremely sparse and weak radar returns of small targets, where residual connections may also preserve noisy local responses. 
Overall, the results show that residual blocks improve the stability and discriminability of graph feature learning, leading to better detection performance in most categories.

% \textbf{Ablation of Virtual Node Layer.} 
% The third row of Table \ref{tab:5} shows the effect of the virtual node layer. Compared to the model with only residual blocks, the inclusion of the virtual node layer improves mAP by 1.3\%. This result demonstrates that the virtual node layer introduces global context and graph identity information into the graph representation, thereby enhancing the model's detection capabilities. Furthermore, we implemented a version that inserts virtual node layer between every convolutional layer, as shown in Fig. 7. In this design, after each convolutional layer output, the features of all nodes are first aggregated into a graph-level representation. This representation is then updated through linear projection and a multi-layer perceptron before being fed back as global context into all node features. This design aims to enable global information exchange at each layer. However, in actual training, this strategy makes it difficult for the bounding box detection head to learn correct bounding boxes, resulting in inadequate detection performance. Therefore, quantitative results for this aspect are not presented in Table 5. Instead, we record the training loss for comparison. As shown in Fig. 6, the purple line represents the training loss for this experiment. It can be seen that the model has convergence problems when the virtual node layer is added before all layers. This phenomenon also indirectly confirms Xing's~\cite{xing2024less} conclusion that over-globalizing can also weaken the expressiveness of graph neural networks.

\textbf{Ablation of Virtual Node Layer.}
As shown in Table~\ref{tab:5}, adding the virtual node layer further improves the mAP from 0.578 to 0.591 on top of residual blocks, with clear gains on Pedestrian, Pedestrian Group, and Large Vehicle. 
This indicates that the virtual node layer introduces useful graph-level context, helping each radar point access global structural information beyond its local neighborhood. 
Such global information is especially beneficial for sparse radar point clouds, where local observations may be incomplete or fragmented. 
However, the AP of Car decreases from 0.727 to 0.696, suggesting that excessive global aggregation may weaken fine-grained local patterns for classes with relatively stable and dense radar reflections. 
This also explains why inserting virtual nodes before every convolutional layer can cause convergence problems.
That is, overly frequent global information injection may lead to over-globalization and reduce the discriminability of node representations~\cite{xing2024less}. Overall, using a single virtual node layer provides a better balance between global context modeling and local feature preservation.

\subsection{Post Process Experiments}
% In order to better apply IRGNN to practical scenarios, we conducted experiments and improvements in the post processing stage. 
% To simulate a real deployment scenario, we conducted experiments on an RTX 2060 12GB GPU. 
% All experiments are carried out under translation invariance.
% We divide the post processing into the following stages: the raw box generation stage, the extraction and filtering stage, and the NMS stage. As shown in Table 6 that the time cost in both the raw box generation stage and the extraction and filtering stage is relatively high. In the extraction stage, the prediction labels, confidence scores, and background scores need to be extracted, which involves a significant number of maximum value and index-of-maximum operations. By vectorizing and optimizing these steps, we achieved batch computation, reducing the time cost of this stage to 22.97\% of the baseline. Additionally, the baseline model had a high VRAM usage during inference. By releasing intermediate states promptly during inference, we reduced VRAM usage to 37.61\% of the baseline, while also decreasing the time cost of the raw box generation stage. Ultimately, we achieved a 38.69\% reduction in time cost and a 62.39\% reduction in VRAM usage, enhancing the deployability of the proposed model.
To evaluate the deployment efficiency of IRGNN, we conduct post-processing experiments on an RTX 2060 12GB GPU under the translation-invariant setting. 
As shown in Table~\ref{tab:6}, the baseline pipeline is mainly bottlenecked by raw box generation and extraction. 
The extraction stage involves repeated operations for prediction labels, confidence scores, and background probabilities.
After vectorization, time cost is reduced from 131.43s to 30.19s, showing the efficiency of batch computation over point-wise processing. 
Since the major VRAM consumption comes from intermediate tensors during raw box generation, further releasing intermediate states reduces VRAM usage from 8.11GB to 3.05GB and also lowers raw box generation time from 224.23s to 176.28s. 
% The NMS cost remains relatively stable, indicating that it is not the main bottleneck.
Overall, the optimized post-processing pipeline reduces total time cost by 38.69\% and VRAM usage by 62.39\%, improving the deployability on resource-limited hardware.

\begin{table}[htbp]
\centering
\caption{Comparisons of post-process time cost and VRAM usage. Time cost refers to the total inference time for the entire validation set. VRAM usage represents the consumption during the raw bounding box generation stage.}
% Comparison of the post processing time cost and VRAM usage before and after the improvements in each stage. Time cost refers to the total inference time for the entire validation set. VRAM usage represents the consumption during the raw bounding box generation stage.
\label{tab:6}
\setlength{\tabcolsep}{8pt}
\begin{tabular}{lcccc}
\toprule
% & \multirow{2}{*}{\makecell{Virtual Node\\Layer}} 
\multirow{2}{*}{Model} 
& \multirow{2}{*}{\makecell{Raw Box\\Cost (s)}} 
& \multirow{2}{*}{\makecell{Extraction\\Cost (s)}} 
& \multirow{2}{*}{\makecell{NMS\\Cost (s)}} 
& \multirow{2}{*}{\makecell{VRAM\\Usage (GB)}} \\
 & & & & \\
% Model & Raw Box Cost/s & Extraction Cost/s & NMS Cost/s & VRAM Usage/GB \\
\midrule
Baseline & 224.23 & 131.43 & 21.61 & 8.11 \\
+Vectorization & 225.34 & 30.19 & 20.94 & 8.10 \\
+Intermediate Release & 176.28 & 32.29 & 22.73 & 3.05 \\
\bottomrule
\end{tabular}
\end{table}

\section{Conclusion}
In this study, we propose \textbf{IRGNN}, an invariant graph neural network framework for radar point cloud object detection in autonomous driving. IRGNN constructs invariant graph representations from sparse radar points and employs an improved MPNN with residual connections and a virtual node layer to enhance feature propagation, reduce over-smoothing, and capture global context. Experiments on RadarScenes demonstrate that IRGNN achieves competitive performance against existing radar-based methods.

% bibtex
\bibliographystyle{splncs04}
\bibliography{referencesflie}
\end{document}